\documentclass{article}

\usepackage[a4paper]{geometry} 

\usepackage{amsmath}
\usepackage{graphicx}
\usepackage[utf8]{inputenc}
\usepackage[backend=biber, 
citestyle=authoryear, style=authoryear,
maxcitenames=2, sortcites=true,
doi=false,url=false,isbn=false,dashed=false,maxbibnames=99,uniquename=false, uniquelist=false,
backref=false,
eprint=false
]{biblatex}

\usepackage{outlines}

\title{Quantifying the dynamics of topical fluctuations in language}
\date{\vspace{-5ex}} 
\author{Andres Karjus\textsuperscript{1}, Richard A. Blythe\textsuperscript{1,2}, Simon Kirby\textsuperscript{1}, Kenny Smith\textsuperscript{1}\\ \small \textsuperscript{1} Centre for Language Evolution, School of Philosophy, Psychology and Language Sciences,\\ \small University of Edinburgh;\\ \small \textsuperscript{2}School of Physics and Astronomy, University of Edinburgh \\ \footnotesize \tt a.karjus@sms.ed.ac.uk, \{r.a.blythe, simon.kirby, kenny.smith\}@ed.ac.uk }

\begin{document}
	
	\maketitle{}
	
	\begin{abstract}
		The availability of large diachronic corpora has provided the impetus for a growing body of quantitative research on language evolution and meaning change. 
		The central quantities in this research are token frequencies of linguistic elements in texts, with changes in frequency taken to reflect the popularity or selective fitness of an element.
		However, corpus frequencies may change for a wide variety of reasons, including purely random sampling effects, or because corpora are composed of contemporary media and fiction texts within which the underlying topics ebb and flow with cultural and socio-political trends. 
		
		In this work, we introduce a simple model for controlling for topical fluctuations 
		in corpora---the \emph{topical-cultural advection model}---and demonstrate how it provides a robust baseline of variability in word frequency changes over time. We validate the model on a diachronic corpus spanning two centuries, and a carefully-controlled artificial language change scenario, and then use it to correct for topical fluctuations in historical time series.
		Finally, we use the model to show that the emergence of new words typically corresponds with the rise of a trending topic. This suggests that some lexical innovations occur due to growing communicative need in a subspace of the lexicon, and that the topical-cultural advection model can be used to quantify this.\footnote{
			A previous, considerably shorter version of this paper outlining the basic model appeared as an extended abstract in the proceedings of the Society for Computation in Linguistics \parencite{karjus_topical_2018}.
			Code to run the analyses described in this paper is available at 
			{\scriptsize\url{https://github.com/andreskarjus/topical_cultural_advection_model}}.
		}
		\smallskip
		
		Keywords: advection, lexical dynamics, language change, language evolution, frequency, topic modeling, corpus-based
	\end{abstract}
	
	\section{Introduction} \label{sec:Introduction}
	
	Elements of a language, be they words or syntactic constructions, never exist by themselves, but in some context. Contexts, or topics, tend to change with the times, along with the world that they describe. These changes are expected to be reflected in (representative, balanced) diachronic corpora. If a particular topic---be it computers, cuisine or terrorism---rises or falls in public interest or newsworthiness, it would be reasonable to expect a similar effect in the corpus frequencies of lexical elements relevant to the given topic, particularly content words such as nouns\footnote{
		We will use the terms `word', `lexical item', `linguistic variant' and `linguistic element' more or less interchangeably in the following text, depending on the literature or subfield being discussed.
	}.
	It follows from this that the changing popularity of some words, apparent from raw corpus frequencies, might well be explained simply by the rise or fall of their most prevalent topics, rather than being a product of other aspects driving language change, such as sociolinguistic prestige or inherent contextual fitness. 
	
	This paper seeks to investigate this idea, which we believe is rather intuitive and widely held, yet to our knowledge has not been formalized in a quantitative way. We will argue that by doing so, we arrive at an informative baseline for frequency-based approaches to lexical dynamics and language change in general. 
	In particular, we show its potential for quantifying topic-driven innovations in the lexicon, and its utility in distinguishing selection-driven change from changes stemming from language-external factors, which manifest as topical fluctuations. 
	
	More precisely, we introduce a quantitative measure of topical change that we call \emph{advection}, a term borrowed from physics where it is used to denote the transport of a substance by the bulk motion of a fluid.
	The analogy is that words are swept along by movements (increases or decreases in frequency) of associated topics. We implement a topical advection measure using a readily interpretable computational technique based on a robust method from distributional semantics. This approach requires very little tuning of global parameters and produces reasonable results given a sufficiently large corpus. As we will show, it is capable of capturing the effect of changing topic frequencies on the frequencies of individual words.
	
	We begin in Section~\ref{twoissues} by providing a brief overview of the state of the art of corpus-based evolutionary language dynamics research and identify the difficulties associated with disentangling different contributions to word frequency changes that may be of interest. We introduce the \emph{topical-cultural advection model} in Section~\ref{modelsection}, and define our measure of advection in terms of the frequency change of words associated with topics. 
	We first show (Section~\ref{results-coha}) that advection is positively correlated with word frequency changes in the Corpus of Historical American English (COHA), 
	indicating that the model successfully captures a component of language change.
	In Section \ref{results-simu} we test the advection model by showing that it correctly associates word frequency changes with a stylistic shift in an artificially-constructed corpus. We then show
	how it can be used to adjust frequency time series (Section \ref{results-decomposition}), and finally (Section \ref{results-innovation}) how it also allows us to quantify the propensity for new words to emerge alongside trending topics.
	
	We conclude that topical advection should be controlled for in any corpus-based research which relies on the (changing) frequencies of lexical items to make claims about patterns or mechanisms of language change. While this paper focuses on language, we believe that the same basic approach could also be utilized in studying the rise and fall of other products of human culture, given appropriate databases or corpora.

	\section{Background: corpus-based approaches to lexical dynamics and language evolution}\label{twoissues}

	A question that often arises in corpus-based evolutionary language dynamics is the causal origin of language change. A key difficulty lies in disentangling the many different possible causes of language change, some of which may be of greater or lesser interest. 
	A number of factors operating on the level of the individual speaker that potentially influence linguistic selection have been proposed and tested, either in experimental settings, simulations, or corpora with speaker metadata --- such as
	the competing pressures of learnability, expressivity, simplicity and efficiency \parencites{kirby_cumulative_2008,smith_linguistic_2013,carr_cultural_2017,kanwal_zipfs_2017,zipf_human_1949,enfield_transmission_2014,culbertson_simplicity_2016},
	egocentricity and content biases \parencite{tamariz_cultural_2014},
	socially conditioned variation \parencite{samara_acquiring_2017},
	and various other social effects \parencites{calude_modelling_2017,lev-ari_experimental_2014,labov_principles_2011}. 
	While language change is perpetuated by the utterance selections of individual speakers over time, %
	some factors also influencing selection may be seen as properties of the population, or those of the linguistic system, such as
	various structural-phonological properties \parencites[e.g.][]{szmrecsanyi_about_2016,ohala_origin_1983}, 
	phonological dispersion and clustering \parencites{dautriche_wordform_2016,dautriche_words_2017,newberry_detecting_2017},
	polysemy \parencites{hamilton_diachronic_2016,calude_modelling_2017}, social network properties
	\parencite{baxter_modeling_2009,castello_agent-based_2013}, top-down language regulation \parencites{daoust_language_2017, ghanbarnejad_extracting_2014,rubin_language_1977,amato_dynamics_2018}, community consensus and relative prestige associated with different variants and languages \parencites[cf.][]{pierrehumbert_model_2014,abrams_modelling_2003-2,hernandez-campoy_handbook_2012,labov_principles_2011}.
	However, some changes may be a result of purely random effects, as individual speakers have access only to a finite sample of utterances (cf. Section \ref{drift}).
	
	In evolutionary terms, this amounts to the problem of teasing apart drift from selection in language change. Even where one can identify a systematic component to a change (selection), factors that might be of interest from a linguistic perspective 
	need to be disentangled from those that are driven by changes in society and culture, or appear due to uneven sampling of genres, registers or topics in a corpus
	\parencite{szmrecsanyi_about_2016,szmrecsanyi_culturally_2014,hinrichs_which-hunting_2015,pechenick_characterizing_2015}.
	Such considerations have come to the fore due to sharp increases in the availability of quantitative data over the last decades. These datasets record how languages are used (corpora), what their distinguishing features are (typological databases) and to what extent languages are used (demographic databases). This development has given rise to the field of \textit{language dynamics}, which has been described as an interdisciplinary approach to language change, evolution, and interlanguage competition, relying on large databases and quantitative modeling, including simulation-based approaches \parencite{wichmann_emerging_2008}. %
	Since our contribution applies to corpus research first and foremost, our focus in the following brief review will be on this strand of language dynamics.

	\subsection{Previous research}  %
	
	Large diachronic collections of language use are of greatest utility from the perspective of understanding language change, as from these one can extract trajectories of change and dynamics of competition between communicative variants. One body of research aims to quantify statistical laws of language change over time, those of word growth and decline, and relationships between word frequencies and lexical evolution \parencites{keller_connectivity_2013,keller_word_2014, feltgen_frequency_2017, pagel_frequency_2007,newberry_detecting_2017,lieberman_quantifying_2007,cuskley_internal_2014,amato_dynamics_2018}. 
	This has also involved claims regarding the effects of real-world events (like wars) on these processes \parencite{wijaya_understanding_2011, petersen_statistical_2012, bochkarev_universals_2014}.
	
	There is also an emerging strand of research investigating semantic change and language dynamics from the point of view of meaning, using diachronic corpora and distributional semantics methods. These include the various flavors of Latent Semantic Analysis \parencite{deerwester_indexing_1990} and word2vec \parencite{mikolov_distributed_2013}. This research broadly falls into two categories: methods proposals usually accompanied by exploratory results \parencites{sagi_tracing_2011,gulordava_distributional_2011,wijaya_understanding_2011,jatowt_framework_2014,kulkarni_statistically_2015, hamilton_diachronic_2016,frermann_bayesian_2016,schlechtweg_german_2017,dubossarsky_outta_2017,kim_temporal_2014,rosenfeld_deep_2018} --- and applications of such methods, usually with more specific linguistic questions in mind \parencites{hamilton_cultural_2016,xu_computational_2015,perek_using_2016,rodda_panta_2016,dubossarsky_verbs_2016,dautriche_wordform_2016}. Notably, all of these approaches are, one way or another, based on (co-occurrence) frequencies of words, and as such are naturally subject to sampling biases potentially introduced by uneven representation of topics and genres in a corpus.
	
	We believe our contribution is also relevant for traditional corpus linguistics, or research more geared towards investigating specific phenomena in some target language(s)---if it involves counting frequencies of words or other elements of speech in diachronic corpora, and using these counts in explanatory models. In all of these cases, it is necessary to deal with factors that serve to confound the explanatory factor of interest, for example, those that are specifically linguistic,
	such as various language processing and transmission biases.
	In particular, as noted above, there is a need to separate random and systematic effects, and frequency changes arising from changes in topic and genre across the corpus and over time. We expand on both confounds below.

	\subsection{Confound 1: language change involves drift} \label{drift}

	It is widely agreed that not all language change is necessarily caused by selection by speakers for certain variants or utterances, but also involves random processes (i.e., drift, or neutral evolution) \parencite{sapir_language._1921,hamilton_cultural_2016,blythe_neutral_2012,newberry_detecting_2017,jespersen_language_1922,reali_words_2010,andersen_structure_1987}.
	Naturally, this should be taken into account in a diachronic study of language. This requires some way of distinguishing changes resulting from drift and those, potentially more interesting ones, resulting from selection. 
	
	Our proposal is by no means the first attempt to construct some form of baseline or null model against which potential cases of directed change can be compared. There have been various proposals to carry over the selection and neutral drift paradigm from evolutionary biology, where drift refers to cases for differential replication without selection (cf. \cite{croft_explaining_2000}). It has been argued that a prerequisite for studying language change through this paradigm would be the construction of well-informed null models \parencite{blythe_neutral_2012}. Proposals in this vein tend to rely directly on or draw from Kimura's neutral model of evolution and the Wright-Fisher model \parencites{kimura_population_1994, ewens_mathematical_2004}. Alleles are equated with linguistic variants and neutral evolution (drift) with (neutral, random) language change \parencite{reali_words_2010}. 

	Adopting this framework, \textcite{newberry_detecting_2017} apply tests developed in genetics for distinguishing drift and selection to frequency time series of competing linguistic variants. In particular, they apply the Frequency Increment Test \parencite{feder_identifying_2014}, and do so on three test cases of changes in the grammar of the English language. They conclude that this constitutes a systematic approach for distinguishing changes likely resulting from linguistic selection rather than drift \parencite[however, cf.][for a reanalysis]{karjus_challenges_2018}.  
	With the culturomics proposal \parencite{michel_quantitative_2011} in mind, \textcite{sindi_culturomics_2016} propose another model to detect departures from neutral evolution in word frequency variation, based on comparing frequency series with randomly generated baselines.
	
	In a slightly different sense, the notion of `(linguistic) drift' has also been used previously in a computational semantics study \parencite{hamilton_cultural_2016}. Drift is defined there as semantic change stemming from (presumably regularly ongoing) change in language---not a reflection of considerable change in the culture that a particular language codifies. The latter is labeled as `cultural shift', which is claimed to be more common in nouns than verbs.
	Detecting `significant' changes in word meaning has also been attempted \parencite{kulkarni_statistically_2015}, with the two aforementioned approaches using a similar distributional semantics method for determining semantic similarity across time, and the latter employing a similar significance detection method as \textcite{feder_identifying_2014}.
	
	The concept of linguistic drift is also commonly utilized in computational modeling of experimental communication data, where the null model, without communicative biases \parencite[such as bias for egocentric coordination or superior expression, cf.][]{tamariz_cultural_2014} would consist of randomized changes, or drift. The question of distinguishing selection from drift has also arisen more widely in cultural evolution, for example, in the contexts of prehistoric pottery \parencite{crema_revealing_2016}, keywords in academic publishing \parencite{bentley_random_2008} and baby names \parencite{hahn_drift_2003}.
	
	Another take on neutral evolution was proposed by \textcite{stadler_momentum_2016}, who demonstrated using a simulation model that language change may also self-actuate without selection but via momentum, whereby variants simply become more popular by virtue of having gradually become more popular. This model produces S-shaped frequency change curves, which have been argued to be a characteristic of language change \parencite{blythe_s-curves_2012}. Relatedly, a similar S-shaped trajectory was seen in a model where a neutral process of language acquisition interacts with a dynamic social network structure \parencite{kauhanen_neutral_2017}   %

	\subsection{Confound 2: language is not independent of its environment}\label{problem2}
	
	No linguistic element exists in isolation: we use language to communicate about salient events in the world, and the language in use in a given time period therefore indirectly reflects the events, concerns and preoccupations of that time.
	These reflections should be observable in a representative corpus.
	The potential effect of real-world changes and hot media topics on corpus-based language usage patterns have been noted in multiple recent studies (see below). However, the way this is approached varies between studies with different aims. We observe at least three ways the connection between language use and real-world change has been considered: 
	as a minor by-product of corpora;
	as an assumption for language-based culture research; and thirdly, 
	as a factor to be necessarily accounted for in linguistic analysis. All of these deserve further discussion.

	\subsubsection{Topical-cultural impact on corpora as an inconsequentiality} %
	
	In a study of mathematical approaches to detecting selection (against drift, cf.\ Section~\ref{drift}) \textcite{sindi_culturomics_2016} observe that words with very similar frequency change patterns also qualitatively belong to similar semantic clusters or topics (e.g., words related to war increasing during periods of war at similar rates). 
	Since their focus is on evolutionary selection dynamics, the topical effect is discussed in passing. \textcite{keller_connectivity_2013} look into word formation dynamics and also observe qualitatively that cultural changes seem to be reflected in the dynamics of the larger morpheme families, but do not explore further.
	
	\subsubsection{Topical-cultural impact on corpora as an assumption}
	
	The field of `culturomics' is based on the assumption that changes in the sociocultural environment of a language should be reflected in the concurrent usage of its lexical items.  Word frequencies in large diachronic collections of texts (such as Google Books) are seen as an interesting way of observing and studying historical real-world changes \parencites{michel_quantitative_2011, bentley_books_2014}. It has also been noted that times of change and conflict, such as wars and revolutions, are observable in language dynamics, such as the emergence of new words \parencites{bochkarev_universals_2014, bochkarev_average_2015} and word growth rates \parencite{petersen_statistical_2012}. \textcite{petersen_statistical_2012} conclude that ``[t]opical words in media can display long-term persistence patterns /.../ and can result in a new word having larger fitness than related `out-of-date' words". 
	Socio-political change can in some cases be observed in the contemporary (distributional) semantics of words, e.g., \textit{Kennedy} being associated with \textit{senator} before and \textit{president} after the year of his election \parencite{wijaya_understanding_2011}. 
	There have been at least two claims of correlations between changes in language and political processes (\textcite{frimer_decline_2015} on the US Congress, \textcite{caruana-galizia_politics_2015} on Nazi Germany), although these have both recently been criticized for methodological errors resulting in spurious correlations \parencite{koplenig_why_2017}. 
	The culturomics approach, and research based on the Google Books corpus in particular, has been recently criticized for ignoring important issues such as metadata of the texts underlying the corpus \parencite{koplenig_impact_2017} and unbalanced sampling of topics, genres or authors in corpus composition \parencite{pechenick_characterizing_2015}.
	
	\subsubsection{Topical-cultural impact on corpora as a problem}\label{twoproblems3}
	
	While the relationship between topicality and language use allows us to use language as a window into changes in the world, as claimed by practitioners of culturomics, it poses a problem if we want to use fluctuations in those same patterns of language use as a diagnostic for linguistic, rather than sociocultural, change. In recent years a number of authors have drawn attention to the importance of controlling for contextual factors such as genre and topic, with some voicing the concern that studying language change via corpus frequencies of linguistic elements alone could potentially be very much misleading. We review some of these below.
	
	\textcite{lijffijt_ceecing_2012} are concerned with testing the assumption that a single-genre general purpose corpus should be relatively homogeneous over time. They find that the period of the English Civil War had an identifiable effect on word frequencies in the Corpus of Early English Correspondence, which they attribute to the over-representation of war-related topics and authors with a military background, violating the assumption of homogeneity.
	In a corpus study on the English \textit{which-that} alternation, \textcite{hinrichs_which-hunting_2015} emphasize the importance of controlling for genre and register, since those alternating variants are associated with different genres. 
	In a study on the evolution of the English genitive markers, \textcite{szmrecsanyi_about_2016} --- lamenting the unreliability of corpus frequencies in general --- reasons that while a ``proper" grammatical change has taken place, ``[a] good deal of the diachronic frequency variability in the dataset can be traced back to environmental changes in the textual habitat". They point out that the shifting nature of the topics in the news section of their diachronic English language corpus --- in particular, the coverage of non-animate entities such as collective bodies --- plays a role in the changing frequencies of \emph{of}-genitives, their object of study.
	
	Topical effects have also been suggested to play a role in word survival dynamics and semantic change. In a synchronic sociolinguistic study of M\~{a}ori loanwords in New Zealand English, \textcite{calude_modelling_2017} point out that simple across-corpus loanword frequencies could be misleading in terms of loanword success, since ``certain words and concepts can become more widely used because they might be relevant to certain topics of conversation". 
	Studying the success of loanwords in French news corpora, \textcite{chelsey_predicting_2010} similarly ask if topic matters: is the occurrence of many financial borrowings the result of a high proportion of financial articles in the corpus, or are financial borrowings just more likely to become entrenched? Their conclusion is that, without information on topics, there is simply no way to tell. 
	Investigating the rise and decline of words in online newsgroups, \textcite{altmann_niche_2011} find that while diffusion among users (speakers) is the primary determinant of the success of a word, spread across the conversation threads within newsgroups (which could also be seen as ``topics") also plays a significant role, with both being better predictors than raw frequency. 
	Using a distributional semantics approach, \textcite{rodda_panta_2016} find qualitative support for the idea that the diffusion of Christianity drove semantic change in Ancient Greek, but point to the over-representation of certain genres in their corpus and call for more research on the effects of corpus composition. 
	
	Although many corpora do include metadata on genres and registers, fine-grained topics --- which may well change rapidly within genres like daily news --- are more often than not missing from the picture. 
	Consequentially, there appears to be a widely articulated need across various branches of corpus-based language research for a method to control for topical fluctuations in corpora, as they are recognized to have potentially far-reaching effects on linguistic analyses based on such data, particularly if they make use of frequencies of linguistic elements. The method we introduce below aims to address that issue.

	\section{The topical-cultural advection model}\label{modelsection}
	
	We begin with the simple intuition that if a topic becomes more prevalent, the words describing it, relating to it and possibly giving rise to it, should become more frequent as well. Similarly, the decline of a topic may drive the decline of words related to it. This effect should be clearer for words specific to certain topics, and less pronounced (or absent altogether) for words with a more general meaning. While we do not claim that our approach offers a remedy to all the concerns reviewed above, we will show that it does provide a simple, easily implemented and intuitive baseline for controlling for topic-related effects arising from sociocultural change or uneven sampling of a corpus. 
	In this section we define the topical-cultural advection model. To aid readability, we defer certain technical details of the implementation to a Technical Appendix.
	
	\subsection{Definition of the model}\label{modelsection-definition}
	
	In its simplest form, the topic of a target word in the topical-cultural advection model is defined as the set of words that are most strongly associated with the target word in terms of co-occurrence over a particular period of time. The context sets should be re-evaluated for each period subsample in a corpus, to accommodate for natural semantic change of words (which would also entail changes in context).

	The advection value of a word in time period $t$ is defined as the weighted mean of the changes in frequencies (compared to the previous period) of those associated words. More precisely, the topical advection value for a word $\omega$ %
	at time period $t$ is
	\begin{equation} {\rm advection}(\omega;t) := {\rm weightedMean}\big(  \{ {\rm logChange}(N_i;t) \mid i=1,...m  \}, \, W \big)\end{equation}
	where $N$ is the set of $m$ words associated with the target at time $t$ and $W$ is the set of weights (to be defined below) corresponding to those words. $m$ is a free parameter (we use the value 75 in the following). 
	The weighted mean is simply
	\begin{equation}
	{\rm weightedMean}(X, W) := \frac{\sum x_i w_i }{\sum w_i} 
	\end{equation} 
	where $x_i$ and $w_i$ are the $i^{\rm th}$ elements of the sequences $X$ and $W$ respectively. 
	The log change for period $t$ for each of the associated words $\omega'$ is given by the change in the natural logarithm of its frequencies from the previous to the current period. That is,
	\begin{equation}
	{\rm logChange}(\omega';t) := \ln[f(\omega';t)+s] - \ln[f(\omega';t-1)+s]    
	\end{equation}
	where $f(\omega';t)$ is the number of occurrences of word $\omega'$ in the time period $t$,
	and $s$ is a smoothing constant, to avoid $\log(0)$ appearing in the expression. The value of $s$ is set to 0 if the relevant frequency $f(\omega')>0$, or if both $f(\omega';t)$ and $f(\omega';t-1)$ are zero. Otherwise, $s$ is set to the value equivalent of 1 occurrence after frequency normalization. Simply put, we replace zero-frequencies with small values to be able to compute log frequency change from and to 0. Mentions of log frequencies and log change here and below refer to natural logarithms. See the Appendix for details on why log change is favored over percent change. 

	The crucial ingredient in the model is the set of weights $W$ for the words in $N$. Here, we adopt the positive pointwise mutual information (PPMI) score \parencite{church_word_1990}. We provide details of how PPMI is calculated in the Technical Appendix. The idea is that PPMI assigns a higher score to words that are strongly associated, based on their co-occurrence with other words. 
	While a very general, high frequency word may occur more often in the vicinity of a target word than some specific, low frequency word, the conceptual association between the target and the general word is likely quite low, as the latter co-occurs with many other words as well --- while the topic-specific one likely does not. PPMI captures this notion and downweights co-occurrence counts with such general words. In terms of the advection model, weighting the frequency changes of the context words by their association scores leads to a better model, as context words more strongly associated with the target more likely belong to the same underlying topic.

	\subsection{Connections with previous work}\label{modelsection-previouswork}

	This model builds on the core notions and recent developments in distributional (vector) semantics, where the meanings and topics of words are defined through their vectors of co-occurring words. These vector spaces may be learned directly from data \parencite{mikolov_distributed_2013} or be based on term co-occurrence matrices \parencites{deerwester_indexing_1990,pennington_glove_2014}.
	In all of these approaches, two words with similar vectors (across dimension reduced vector spaces, or across the vocabulary of context words) are considered to have similar meaning. A common measure of similarity is the cosine of the angle between the two vectors. Recently, an alternative has been proposed in the form of the APSyn measure  \parencite{santus_testing_2016}, which involves comparing the rankings of the topmost associated context words 
	instead of the whole vocabulary. The intuition behind APSyn is that only the most associated context words hold relevant information about the target word, while most of the words are likely irrelevant. \textcite{santus_testing_2016} demonstrate the capacity of APSyn to perform as well, and in some cases better than the vector cosine. 
	Considering only top ranking contexts is also similar to \textcite{hamilton_cultural_2016}, who use cosine similarity between word vectors between time periods to measure semantic change, but as a second measure, the extent of the change in a word's similarity to its top nearest neighbors \parencite{hamilton_cultural_2016}. 
	We adopt this approach of considering only the top most $m$ associated context words here to determine a ``topic'' for each word, using PPMI as the association score. 
	
	It is nevertheless worthwhile to compare our PPMI-weighted approach with a more traditional topic model. To this end, we also implemented the advection measure using Latent Dirichlet Allocation (LDA) \parencite{blei_latent_2003}. In this approach, each of its latent $k$ topics (we used $k=500$) is assigned a frequency change value based on the frequency changes in the vocabulary, weighted by their association with the topic (as a latent topic is essentially a distribution across the vocabulary). The topical advection value of a target word is then the mean of the changes in the topic frequencies, weighted according to the probability a word belongs to each given topic. The details of this calculation are given in the Technical Appendix. 
	
	As will be seen below in Section~\ref{results-coha}, the descriptive power of the two models is rather similar. While LDA is widely used, we feel that our simple PPMI-weighted model has certain advantages. In addition to requiring the setting of only a single parameter, it is much less computationally complex (thus faster), and the results are easily interpretable. Specifically, each ``topic'' of a target is a short list of top context words (meaning the advection value, being the weighted mean of their log frequency change values, is on the same scale as the target word log frequency change values). It is also straightforward to observe the behavior of a target word's topic and calculate its advection value both before and after it has entered the language or gone out of use --- by re-using the context word list and the corresponding weights from a period where the target word was already (or still) frequent enough for its topic to be inferred.\footnote{
		Similar extensions for evaluating topics over time exist for the latent topic modeling approach, \parencite[cf.][]{wang_topics_2006-1,blei_dynamic_2006,roberts_structural_2013}, which we will be not examining in further detail here. Furthermore, \textcite{frermann_bayesian_2016} use a Bayesian approach in some aspects similar to classical topic modeling to measure semantic change in a word as change in its distribution of ``contexts" (topics). Their model however appears very demanding in terms of the size of the training corpus.
	}
	
	\section{Results of applying the advection model in a number of language change scenarios}\label{results}
	
	We now turn to two large, representative, POS-tagged corpora, in order to get a sense of how well the topical-cultural advection model performs, and proceed to demonstrate a number of useful applications. We preface the results with a few crucial technical details that apply to all the following subsections, and both the PPMI and LDA based models, while leaving a more thorough description of the parameterizations of the models and relevant corpus preprocessing steps to the Technical Appendix. 
	
	The word counts for each time period (segment) in a corpus were normalized as frequencies per million words (pmw).
	Since cultural effects are likely the most pronounced on content words, particularly nouns \parencite[see also][]{hamilton_cultural_2016}, we only consider common noun targets in the following analyses. For the context vectors (see Section \ref{modelsection-definition}), we exclude stop words and use only content words (based on POS tags). We use the top $m=75$ context words for the PPMI based model. We set a (rather conservative) threshold of a minimum of $100$ occurrences per period for words to be included in the model. If a word occurs less than $100$ times in a corpus period, it will not be assigned a context vector --- thus also no advection value for this period --- nor will it be used as a context word. This comes down to a classical statistical sampling problem: if a word only occurs a few times, then its context vector (topic) is more likely to be composed of quite random words, in a random ranking, while if a word is observed numerous times, the ranking of its (recurring) context words becomes more reliable. %

	This however also means that it is not possible to calculate the advection value for low frequency words like recent innovations and words going out of usage. Since these correspond to periods of particular interest for such words, we experimented with using a `smoothing' procedure %
	to improve the informativeness of the topics.
	Specifically, the `smoothed' data, used for deriving the topics, comprises text from a target period and its preceding period (word counts still correspond to the frequencies in the target period). This procedure increases the chance of inclusion for relevant context words that would otherwise not be present due to being too low frequency in one or both of the periods. Consequently, it also improves the precision of the advection measure for words decreasing in frequency in a given target period.

	\subsection{Topical advection and diachronic language change}\label{results-coha}

	We use the Corpus of Historical American English (COHA) \parencite{davies_corpus_2010} as a test set in order to evaluate the extent to which the model is capable of accounting for variance in word frequency changes. The COHA spans two centuries, starting with 1810, is binned into decade-length subcorpora by default, and is meant to be balanced across genres for each period (news, magazines, fiction, non-fiction; but see the Appendix for details). 
	
	With 20 decades, there are potentially 19 frequency change points that can be calculated for each target word. 
	There are 7551 unique words in the no-smoothing condition, and 75653 data points. There are 10060 words (107475 data points) in the smoothing condition (concatenated data results in more words being above the minimal threshold to be eligible for the advection calculation).
	
\begin{figure}[tbh]
		\noindent
		\includegraphics[width=\columnwidth]{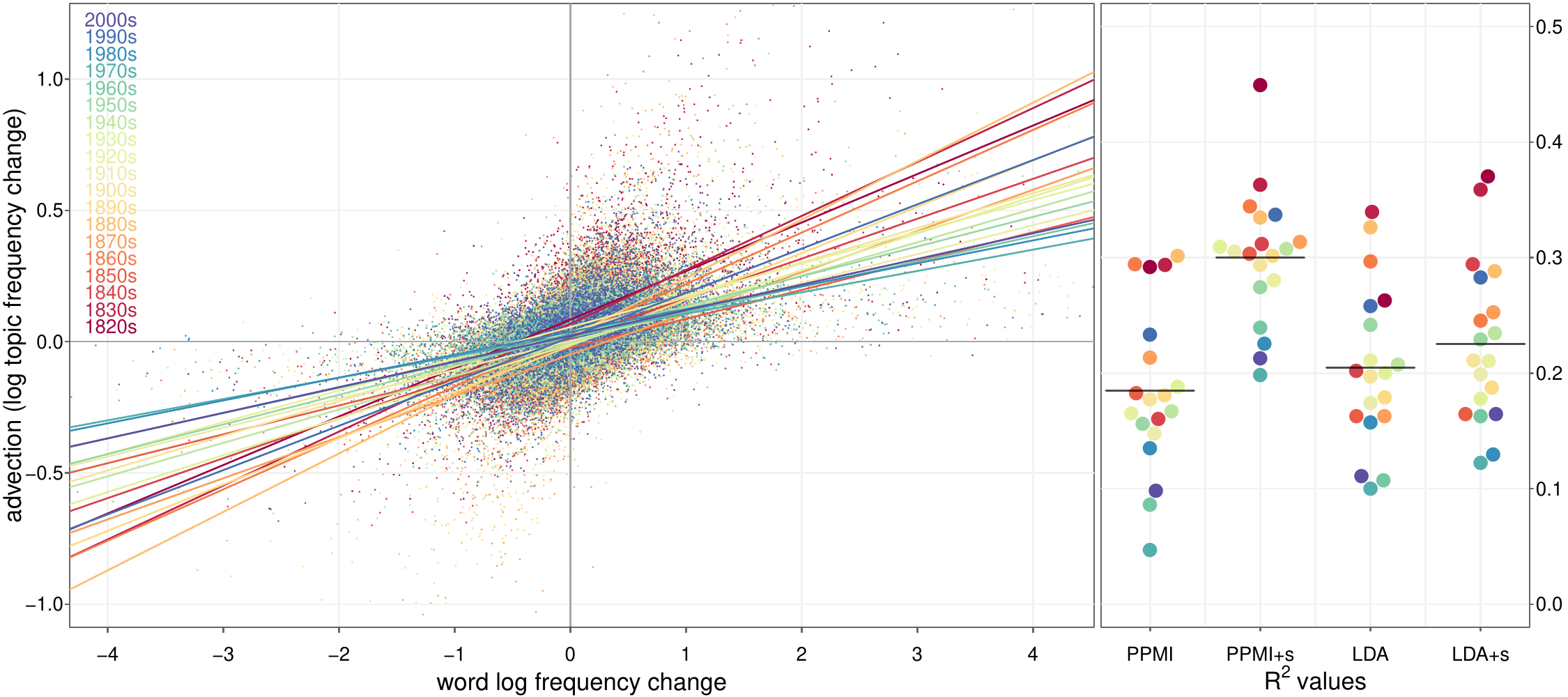}
		\caption{Left panel: log frequency changes of nouns and their corresponding topical advection (log topic change) values from two centuries of language change (from the PPMI-based model with topic smoothing). Each of the 107475 dots indicates the frequency change and advection value of one of the 10060 nouns, colored by decade. As such, many words occur multiple times in this figure. Positive values indicate increase, negative ones indicate decrease. \newline
			Right: $R^2$ values for correlations for each decade. \emph{+s} indicates models with topical smoothing; the black bars mark the means. The PPMI-based models with smoothing have the highest mean $R^2$ of $0.25$. All $p<0.001$.
			This figure illustrates the robust correlation between frequency change and advection. We will be using the same colors to indicate decade subcorpora throughout this paper.
		}\label{fig:scatter_with_examples}
	\end{figure}

	To test the descriptive power of the two aforementioned implementations of the advection model, PPMI-based and LDA-based, we correlate the log frequency change values of common nouns between successive decades in the COHA corpus to their respective advection values (their log topic frequency change values in the same decades).\footnote{
		Importantly, we are not correlating absolute frequencies of words with the absolute frequencies of topics, which could easily lead to spurious correlations (cf. \textcite{koplenig_population_2016} for recent criticisms). 
	}
	The results are presented in Fig.~\ref{fig:scatter_with_examples}. 
	The different scales on the axes indicate that words experience more rapid changes in either direction than topics, as one might expect, topic values being averages of context word frequency changes.

	We find that, as expected, frequency changes correlate significantly and positively with advection, and that the smoothing operation further improves the correlation. The LDA-based and the PPMI-based models yield similar results. The less complex PPMI-based model (with smoothing) performs even slightly better, describing an average of 30\% of variation in noun frequency changes between decades. 
	There is also some variation between decades. The stronger correlations in some decades may be an indication of either a  change in discourse in American English, as chronicled in the corpus, or differences in topical sampling between the subcorpora. 
	We find that the strength of this relationship is in turn positively --- but only moderately --- correlated with observed divergences between distributions of genres in the decade subcorpora (see the Appendix for more details). In short, the advection model tends to describe more variance in word frequency changes between decade pairs which exhibit a larger divergence in their genre distribution (which can be expected to affect the underlying topic distribution).
	
	These results clearly show that topical fluctuations can be expected to explain a significant amount of variability in the change in word frequencies, which one might otherwise be tempted to attribute to other processes, such as selection.  
	As such, the topical-cultural advection measure serves as a useful baseline in any quantitative model predicting frequency changes in linguistic elements.

	\subsection{Artificially-constructed language change based on genres in a synchronic corpus}\label{results-simu}
	
	Having established that advection constitutes one (small but significant) contribution to word frequency change in general, we now test whether our model can identify instances where it is the main contribution to change. This is difficult to determine with natural data, as one does not know \emph{a priori} what the drivers of change are (beyond the genre distribution discussed in the previous section).
	To deal with this problem in a more controlled way, we construct an artificial corpus wherein the main component of change between two subcorpora is a known stylistic shift. We should then find that changes in word frequencies are strongly correlated with topics that are more prevalent in one style than the other.
	
	Specifically, we employ the Corpus of Contemporary American English (COCA)\parencite{davies_corpus_2008}, which is the synchronic cousin of COHA. It consists of contemporary American English data from 1990-2012, again labeled by genres.
	However, in contrast to COHA, COCA is large enough that genre subcorpora from even relatively short time segments contain enough data for training the advection model. This allows us to avoid the potential confound of actual diachronic language change.
	We used only data from a short time span (2005-2010) in the academic journals and spoken language (TV and radio transcripts) subcorpora to construct an artificial ``language change" from academic to spoken style and content, by defining the former subcorpus as one ``period" and the latter as the following one.	 %
	
	We then measured the log frequency changes of nouns, as in the previous section, and their respective advection (log topic frequency change) values. 
	Not surprisingly, among the top decreased are words like \textit{subscale}, \textit{coefficient}, \textit{self-efficacy}, \textit{carcinoma}, \textit{pretest}; while words like \textit{tonight's}, \textit{ma'am}, \textit{fiancee}, \textit{everybody}, and \textit{paparazzi} have all increased with the switch in genre.
	Again, the advection measure correlates positively with frequency change, and describes a notable amount of its variability: in our favored PPMI-based model, we find $R^2 = 0.45$ without smoothing and $R^2 =  0.73$ with smoothing applied.\footnote{
		As there are only two `periods', smoothing here refers to concatenating the entire spoken and academic subcorpora for the purposes of estimating the topics of each word.
	}
	This is to say, the advection model appears to successfully pick up on the genre change, reflected in the high (positive) correlation value --- the decrease in academic and increase in spoken style word frequencies corresponding to the fall of the academic and rise of the spoken topics or genres. Importantly from the perspective of validating our model, the $R^2$ values are higher here than in the analysis of COHA. Presumably there are other forces affecting word frequencies in the COHA besides genre divergences and topic fluctuations; at the same time, the (actual) changes between subsequent decades are likely less stark.

	%

	\subsection{Using advection to adjust for topical fluctuations in time series}\label{results-decomposition}

	Having measured the descriptive power of the advection model and demonstrated how it behaves with re-evaluated topics over time, we now turn to an application of the model to deal with the confounds set out in Section~\ref{twoproblems3}. When it comes to predicting frequency changes of words or any other linguistic elements between periods of time, the advection measure can be included as a control variable in a predictive model (see Section \ref{results-coha}). 
	In the case of time series analysis (i.e., involving multiple changes over time), it is possible to utilize the advection measure as a form of (in the following example, additive) time series decomposition, by carrying out the following operation. For a given word, for every period data point: subtract the advection value (log topic frequency change) of the target word from the log frequency change value of the target word.
	This yields a new series of frequency change values where the topical change component has been removed.
	In this section, we make use of the simple PPMI-based model (with smoothing). The advection values therein are averages over individual word log frequency changes, so the two quantities are on the same natural scale (changes in word frequencies) and can therefore simply be subtracted from each other. 
	See the Appendix for a more technical breakdown of the approach.
	
	The operation described above is similar to seasonal decomposition, a commonly applied approach in (multi-year) time series analysis to control for seasonal ups and downs (e.g., heating costs in cold and warm seasons). In our case, the ``seasonality" (topical fluctuations) is not inferred from the time series itself, but calculated independently. 
	Another way of looking at this is as a way of distinguishing the metaphorical ``word of the day", one that is selected for, from a word that just comes and goes with the ``topic of the day". 
	Adjusting for topics has the potential to be useful in carrying out more objective tests of linguistic selection \parencites[cf.][]{newberry_detecting_2017,sindi_culturomics_2016,bentley_random_2008,blythe_neutral_2012}, by controlling for the topical-cultural element.
	
\begin{figure} %
		\noindent
		\includegraphics[width=\columnwidth]{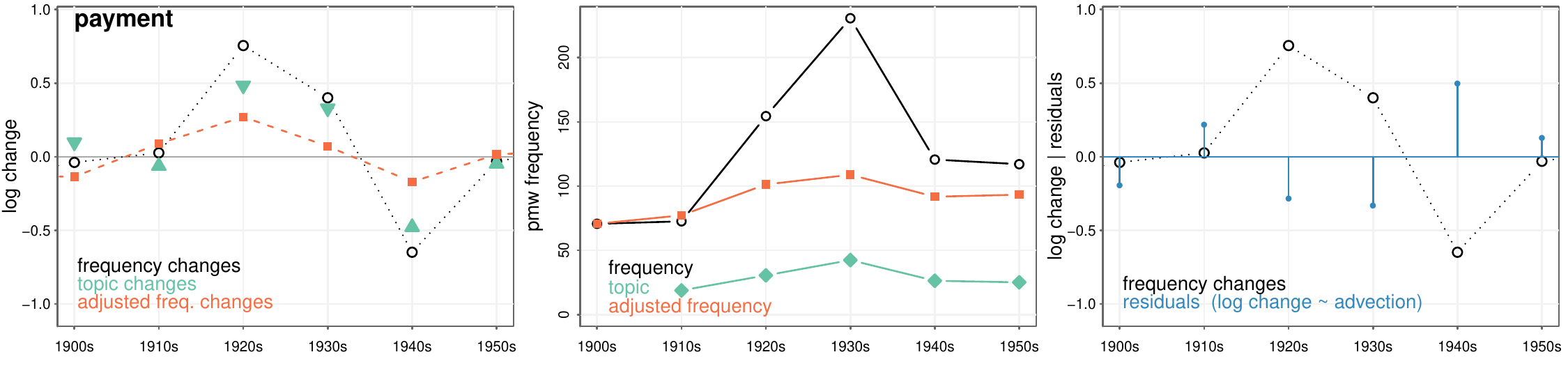}
		\caption{
			Time series of \textit{payment} in the first half of the 20th century. Usage of the word increases considerably in the 1930s, but so does its topic. 
			Black circles: log frequency change values (dotted line), actual frequency (solid line). 
			Green triangles: topic frequency; change values on the left panel, with the triangle pointing up and down corresponding to the adjustment; as relative frequency in the middle panel.
			Orange squares: frequency changes of the word adjusted by subtracting the log topic frequency changes from the word log frequency changes (left; as a reformed series in the middle panel). 
			Note that the green topic line in the middle panel is plotted for reference and only illustrates topic frequency as a relative measure, being a cumulative sum of the log topic changes, initiated with an arbitrary value.
			Blue dots below and above zero on the right side panel: residuals of the target word taken from per-decade regression models. The adjustment operation is generally in line with the residuals: frequency gets adjusted upwards when the residual is positive, and downwards when the residual is negative.
		}\label{fig:timeserieswords4}
	\end{figure}

	Figure~\ref{fig:timeserieswords4} illustrates the results of the adjustment operation on the example of a segment of the time series of the word \textit{payment} in COHA. The left side panel depicts the log frequency changes and the subsequent adjustment. The middle panel shows the same data as actual (per-million) word frequencies. Namely, the time series of word frequencies may be subsequently reformed for visualization purposes, after operating on the change points, as the (exponential of the) cumulative sum of the resulting log change values, initialized with the log frequency of the word at the start of the time series. This however requires selecting the arbitrary initialization value for the cumulative sum, which of course shifts the actual frequency values in the reformed series. The same approach can be used to visualize a topic ``frequency" time series.

	Finally, the right side panel in Fig.~\ref{fig:timeserieswords4} illustrates yet another way of looking at word frequency changes through the lens of advection, making use of regression residuals. We ran a linear regression model for each decade (cf. Fig.~\ref{fig:scatter_with_examples}), where frequency change is predicted by advection. Each blue point above and below the zero line marks the residual value of \textit{payment} in each decade. Above zero indicates that the word is doing better than would be expected by its topic (hinting at selection). Conversely, below zero values indicate that the word is used less than would be expected given the prevalence of its topic.

	One obvious concern with using the advection measure for a decomposition-like operation --- subtracting topic frequency change from word frequency change --- is that it might be over-correcting frequency changes and interfere with observing genuine competition in language, whereby one lexical element is replaced with a synonym over time. To investigate this possibility, we constructed a second artificial corpus, based on 11 decades (1900s-2000s) of the preparsed COHA corpus (cf.\ Section~\ref{results-coha}). The manipulation of the corpus consisted of replacing a set of otherwise stable words with (invented) synonyms in a controlled way. We find that after applying the advection adjustment, the artificially-constructed language change remains untouched, leading us to believe that this adjustment by subtraction does not obscure genuine (although in this case artificial) cases of selection (see the Appendix for a full technical breakdown). 	 %
	

	\subsection{Advection predicts lexical innovation}\label{results-innovation} 

	\textcite[:~194]{mcmahon_understanding_1994} notes that ``new words are most likely to survive, and indeed to be created in the first place, if they are felt to be necessary in the society concerned. This is a difficult notion to formalize, but a well-established one".     %
	Previous empirical research has linked vocabulary size with communicative need as well. Studying color words in 110 languages across the world, \textcite{gibson_color_2017} argue that the communicative needs rising from the environment where these languages are spoken dictates (to an extent) the color naming systems that emerge. In another cross-linguistic study, \textcite{regier_languages_2016} show that the need for efficient communication --- which varies across cultures and environments --- does seem to drive vocabulary size (in their case, of words for `ice' and `snow').
	
	From a historical perspective, this suggests the hypothesis that an increasingly popular topic (i.e. exhibiting positive advection) would be expected to attract new words, providing the detailed vocabulary required---or, conversely, a new word would be expected to exhibit a strong positive advection at its period of first occurrence, compared to the advection values of its topic in previous periods. We are now equipped to test the latter hypothesis.
	
\begin{figure}[tbp]%
		\noindent
		\includegraphics[width=\columnwidth]{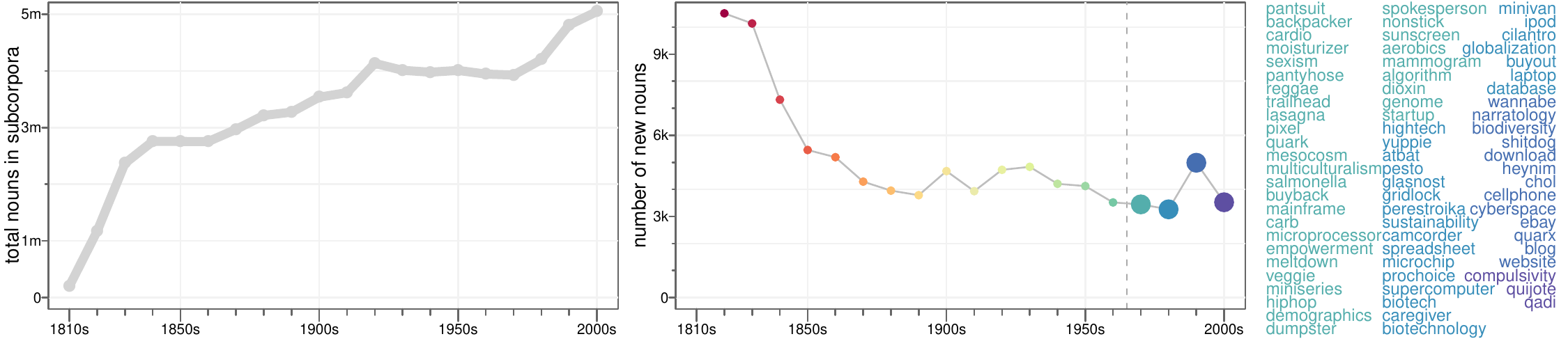}
		\caption{
			Token frequencies of nouns (left) and type frequencies of new nouns (middle panel) in the (preparsed) COHA corpus across period subcorpora. The vertical dashed line on the middle panel indicates the last four decades used to determine the test set of new words in this section; these words are visualized on the right (in corresponding colors).
		}
		\label{fig:nounfreqs}
	\end{figure}

	We identified a test set of 73 ``successful'' novel common nouns from the COHA that meet the following criteria: our successful novel nouns appear as new words in the 1970s to 2000s, and, importantly, occur with high enough total frequency across (at least some of) these decades for their topics to be reliably modeled (it is in this sense that the nouns are ``successful''). Notably, each period of COHA includes a rather large number of new words, but most of them occur at very low frequencies. Figure~\ref{fig:nounfreqs} illustrates the differences in subcorpora sizes across decades in the corpus and the number of new nouns per period.\footnote{
		Note that these counts correspond to our cleaned version of the corpus (cf. Section \ref{results}; this also included the removal of all capitalized words to avoid occurrences of mistagged proper nouns, see the Appendix for details).
		The numbers of ``new" or previously unseen words are likely inflated by the occurrence of spelling mistakes, uncommon words and OCR errors (which commonly end up with the noun tag).
	}
	
	To remedy the small sample problem particularly relevant to new words (that often start out at low frequencies), we again used the simple ``smoothing" technique (see introduction of Section \ref{results}), this time concatenating data from all the last four decades for the purposes of constructing the PPMI-based topic vectors. We chose only novel target words from the last few decades of the corpus in order to carry out the following comparison. 
	
\begin{figure}[thpb]
		\noindent
		\includegraphics[width=\columnwidth]{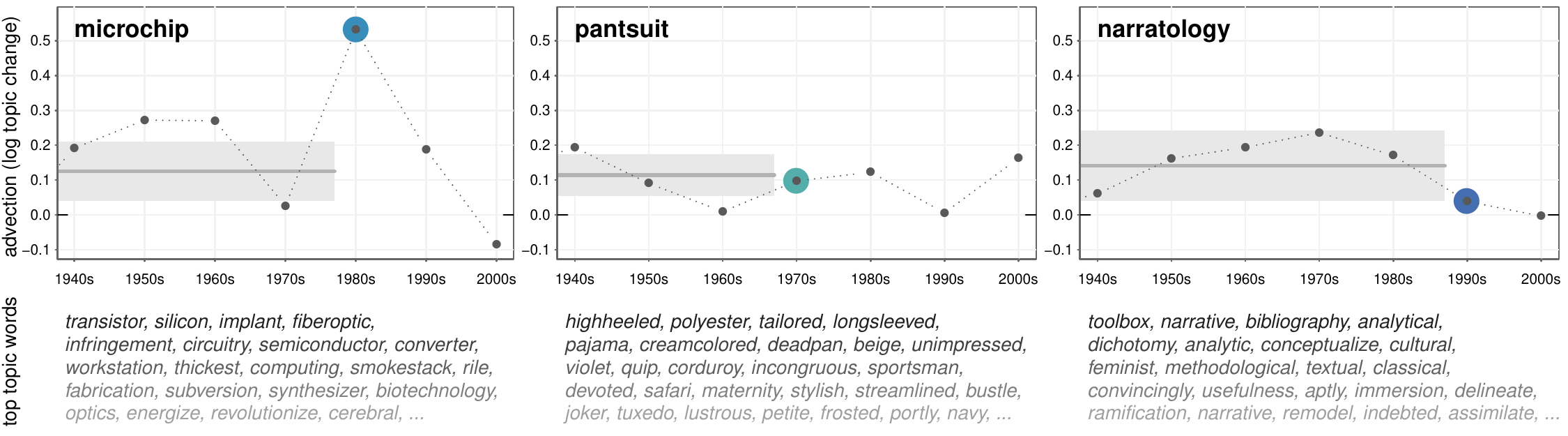}
		\caption{
			Three example novel words.
			The dashed and dotted dark gray line: the advection (log change) values of the topic of the word; above 0 indicates an increase, below 0 a decrease in the topic (note that this is not the frequency of the word, but the mean log changes in the topic).
			The brightly colored circle marks the entry decade of the word --- this is the advection value that is compared against the mean of the preceding advection values. The mean of preceding decades is indicated with the horizontal solid gray line, with a light gray colored confidence interval. The relevant co-occurring topic words are visualized as clouds below each panel (ordered by their PPMI scores).
			\textit{microchip} is among the 58\% of our novel word sample that enter the corpus when its topical advection value is significantly above the mean of the past 10 decades. It is around the mean for \textit{pantsuit}, and below for \textit{narratology}.
		}\label{fig:newwords}
	\end{figure}

	As each topic consists of a list of words, we computed their advection values (log frequency changes) across ten decades preceding the decade where the target word would first occur in the corpus.\footnote{
		Importantly, the advection calculation only took into account words that actually occur (frequency above 0) in a given decade: 0-to-0 frequency changes are not allowed to bias the earlier advection values to be closer to 0. Although some topic words are also new, most topic words do occur in previous decades.
	} 
	In essence, we track how well each topic of each new word is doing throughout a century before the appearance of the innovation. This allows us to measure how many of the (successful) new words belong to topics that exhibit higher advection than before in the period where the new word first appears.
	For 58\% of novel nouns out of the 73, the advection value of the topic associated with the word was found to be above the upper bound of the 95\% confidence interval of the mean of its advection values over the preceding 10 decades (e.g., \textit{microchip}, cf. Fig.~\ref{fig:newwords}). 37\% fell around the means, and only 5\% were below the lower bound of their respective confidence intervals.\footnote{
		We also checked if the large number of new words above their mean advection values could possibly be due to some particular semantic cluster of words that might all belong to a similar (trending) topic and thus inflate the results. We computed the APSyn similarity \parencite{santus_testing_2016} on all pairs of the topic vectors of the 73 nouns and found them to be sufficiently dissimilar.
	}
	
	We also conducted a t-test in the following manner to test the apparent tendency. We calculated the z-score of the advection value of each of the 73 new words at the decade of first occurrence, using the mean and standard deviation values of the previous decades (separately for each of the new words). A one-sample t-test on this set of z-scores indicated that its mean is significantly ($p<0.001$) above zero --- or in other words, the advection values of new words are on average significantly higher at the time of entry than in preceding decades.
	These findings suggests that the appearance of new words does indeed correspond to the rise of certain topics, or the increasing communicative need for new words. Figure~\ref{fig:newwords} illustrates this effect for three novel words that enter into the corpus at different advection values.
	
	\section{Discussion}
	
	A language corpus is essentially a sample of aggregated utterance selections by (a sample from) the population of speakers. 
	In principle, factors which have been claimed to drive selection could therefore be tested for in a corpus, as some have been --- a diachronic one in case of claims about change dynamics, and synchronic if the claims concern properties of language as such. Models connecting individual-level biases and population-level observations have been recently proposed as well \parencite{kandler_inferring_2017,kandler_generative_2018}. In the diachronic case, if the analysis was to involve changing frequencies over time, then the topical-cultural advection model would be straightforwardly applicable as a factor of control or baseline change.
	It could likely also improve tests for selection and drift \parencites[cf.][]{newberry_detecting_2017,sindi_culturomics_2016,bentley_random_2008,blythe_neutral_2012} by adjusting for the component of fluctuating topics presumably driven by socio-cultural processes or ``newsworthiness". While contextual suitability for a topic could be argued to be itself a signal of selection, our model remains applicable, allowing for a quantification of that signal, or to be used as a predictor on its own, as shown in Section \ref{results-innovation}.%
	
	In the case of natural language, our technique for measuring topical advection does require a certain amount of data to be reliable (in terms of inference of the topics, cf. Section \ref{results}). 
	As such, it is directly applicable to (sufficiently large) corpora, regardless of them consisting of newspapers, books, transcripts, dialogs or interviews. This includes both diachronic corpora (i.e., involving 2 or more time periods) and synchronic corpora (consisting of distinct subcorpora, cf. Section \ref{results-simu}). It is less likely to be useful in experimental settings.
	In principle, the advection model could also be used in other domains of cultural evolution, where there is diachronic data available about the systematic co-occurrence of traits or properties (in lieu of context words) of cultural elements (in lieu of target words, such as nouns in the previous sections).
	
	In a sense, our model also orthogonally complements the momentum model of \textcite{stadler_momentum_2016}. They demonstrate, using a simulation of language evolution, that change can self-perpetuate without selection, when a linguistic variant gains enough momentum in its frequency changes over time. While they model momentum from the frequency change of a variant itself, we model the frequency change of a variant potentially driven by the frequency change in its immediate contextual topic (not itself), or what could be called `topical momentum'.

	\section{Conclusions}
	
	We presented the topical-cultural advection model, along with two potential implementations, as a straightforward method capable of capturing topical effects in frequency changes of linguistic elements over time. In particular, we demonstrated that the model 
	accounts for a considerable amount of variability in noun frequency changes between decades in a corpus spanning two centuries, 
	retains its capacity when used on an artificially sampled corpus where a change in style and contents has been simulated,
	and can, to an extent, predict lexical innovation, based on increases in topic frequencies. We also introduced a way of using the advection measure for time series adjustment to distinguish (presumably selection-driven) changes from topical fluctuations (or potentially uneven corpus sampling).
	We conclude that the topical-cultural advection model 
	adds an important analytical approach to the toolkit
	for corpus-based lexical dynamics research, or any investigation drawing inference from changing frequencies of linguistic (or other cultural) elements over time.

	\section*{Acknowledgments}
	
	The authors would like to thank Stella Frank for advice with implementing the Latent Dirichlet Allocation version of the model; Jennifer Culbertson for fruitful discussions. The first author of this research was supported by the scholarship program Kristjan Jaak, funded and managed by the Archimedes Foundation in collaboration with the Ministry of Education and Research of Estonia.

	\section*{Technical appendix}
	
	\subsection*{Notes on preprocessing and parameters}
	
	We take a number of preprocessing steps to ensure a reasonable quality in the inference of the topic vectors that underlie the advection model. 
	Both in the case of COHA and COCA, we exclude stop words (and also a list of known OCR errors) and use only content words (based on corpus POS tags). While COHA and COCA distinguish proper and common nouns in its tagging, we noticed quite a few proper nouns were tagged as common ones, hence we decided to remove all capitalized words (this is particularly relevant in the context of Section \ref{results-innovation}, where we needed to avoid detecting mistagged proper nouns as innovative common nouns). We also reduced variability in spelling by removing hyphens, and replaced all sequences of numbers within content words with a placeholder. 
	
	We used a context window of 10 words on both sides of the target word (after the removal of stop words, etc.), linearly weighted by distance, for inferring co-occurrence. The co-occurrence matrices were subsequently weighted, using the positive pointwise mutual information (PPMI) between each target word $w$ and context word $c$:
	\begin{equation}
	{\rm PPMI}(w, c) := \max\left\{ \log_2 \frac{ P(w,c) }{P(w)P(c)}, 0 \right\}
	\end{equation} 
	
	This is essentially a weighting scheme that gives more weight to co-occurrence values of word pairs that occur together but not so much with other words, and less weight to pairs that co-occur with everything. Since we set a threshold of 100 occurrences per period for a word to be included, we circumvent the known small values bias of PPMI. Since we use positive PMI, all co-occurrence values end up as $\geq0$. See e.g. the textbook by \textcite{jurafsky_speech_2009} for further details and examples.
	
	For the advection model based on vectors drawn from a PPMI-weighted co-occurrence matrix, we use the top $m=75$ context words as the topic (having observed that very small values lead to less reliable topics, while considerably larger values deteriorate the results in some cases). Importantly, the word counts (that underlie the log change values, which in turn make up the advection values) for each period were normalized to per million frequencies using the total word count in that period (periods corresponding to decades by default in COHA).
	
	\subsection*{Algorithmic description of the topical-cultural advection model}
	
	\begin{outline}[enumerate]
		\1 Preprocessing steps
		\2 (optional) Basic text cleaning (using a list of OCR errors, a list of stop and function word tags, words shorter than 3 characters), keep only content words; remove all capitalized words to avoid proper nouns
		\2 (optional) Affix tags to words in the POS class of interest (e.g., nouns in our case; more tags and more specific tags improve disambiguation, but also increases sparsity)   
		\2 Split texts in the corpus files according to document delimiter tags (e.g., `\#\#' in COHA) to avoid word co-occurrence windows crossing document boundaries
		\2 Aggregate and store the preprocessed texts according to chosen periods (e.g., decades)
		\1 Calculate frequency change
		\2 Count the frequencies of words in each period subcorpus and normalize the counts to obtain comparable (relative) values (subcorpora may be of different size)
		\2 For each word $\omega$, between each pair of successive time periods $t$, calculate the log frequency change value: ${\rm logChange}(\omega;t) = \ln[f(\omega;t)+s]-\ln[f(\omega;t-1)+s]$ where $f(\omega;t)$ is the number of times word $\omega$ appears in the corpus during time period $t$. Note we use the $+s$ offset to avoid $\ln(0)$, and set the value of $s$ to the equivalent the value corresponding to 1 occurrence after normalizing to per-million counts. $s$ is set to 0 if $f(\omega)>0$ or if both frequencies are 0.	 %
		\1 (A) Topics and advection (if using the PPMI vectors based approach)
		\2 Generate term co-occurrence matrices for each period (e.g., target words as rows and context words as columns), using a context window of some length (we used $\pm10$, and linearly weighted context words by distance within the window)
		\3 (optional) If targeting a specific POS class, filter the matrices by keeping only rows with the previously affixed tag
		\3 (optional) Filter by setting a frequency threshold for a word to be included (we used a threshold of 100 raw occurrences per period (or per concatenated dataset, if using smoothing)
		\2 Apply positive pointwise mutual information (PPMI) weighting to each matrix%
		\2 Retrieve and store relevant context words for each target, in each period (i.e., sort each row of each matrix and store the top $m$ context words, along with their PPMI weights in that row; we used $m=75$)
		\2 (optional) to apply the ``smoothing'' operation, concatenate data from pairs of successive periods instead, and apply the previous 3 steps
		\2 For each target word $\omega$, in each period $t$, calculate its advection value:
		\3 The advection values is a weighted mean over the log frequency change values in the set (of length $m$) of a target's context words $N$ (i.e., the `topic'), with their PPMI values as the weights $W$; ${\rm advection}(\omega; t) := {\rm weightedMean}\big(  \{ {\rm logChange}(N_{i};t) \mid i=1,\ldots,m  \}, \, W \big)$, where ${\rm weightedMean}(X, W) := \frac{\sum_i x_i w_i}{\sum_i w_i}$
		\setcounter{enumi}{2}%
		\1 (B) Topics and advection  (if using the LDA topics based approach)
		\2 Train Latent Dirichlet Allocation \parencite{blei_latent_2003} models for all period subcorpora (we used the following parameters: $\alpha = \beta=0.1$, $k=500$, maximum allowed iterations: 5000)
		\2 For each word $\omega$ in each period $t$, calculate its advection value:
		\3 Given the $k$ topics, $\tau$, identified by LDA, we determine the number of times $n(\omega,\tau)$ that each word $\omega$ appears in each of the topics $\tau$. From this we can define the two conditional distributions $p(\omega|\tau) = n(\omega,\tau)/\sum_{\omega'} n(\omega',\tau)$ and $p(\tau|\omega) = n(\omega,\tau)/\sum_{\tau'} n(\omega,\tau')$. Given a word frequency change ${\rm logChange}(\omega;t)$ at time $t$,  its contribution to the change of the topic $\tau$ is ${\rm logChange}(\omega;t)p(\tau|\omega)$. 
		
		To construct the advection of a target word $\omega$, we need to determine the frequency changes of all topics that are coming from words other than $\omega$, i.e., ${\rm logTopicChange(\tau; \omega, t)} = \sum_{\omega'\ne\omega} p(\omega'| \tau) {\rm logChange}(\omega';t)p(\tau|\omega') / [1- p(\omega|\tau)]$. Then, ${\rm advection}(\omega;t) = \sum_{\tau} {\rm logTopicChange(\tau; \omega, t)} p(\tau|\omega)$. The last part is thus analogous to point 3.5.1, the change in topic frequency being operationalized as a weighted mean of the changes in word frequencies, with weight from the distribution of words over topics.
		
		\1 (optional) Measure the descriptive power of the advection model by correlating the advection value of each word in each period to its respective log frequency change value.	
	\end{outline}

	\subsection*{Additional remarks on the model and data processing}
	\subsubsection*{For our purposes, logarithmic change is more useful than percentage change}

	We opt to quantify the changes in word counts between different time period subcorpora, using the measure of logarithmic difference --- thus referring to it simply as `log change' \parencites[cf. also][]{altmann_niche_2011,petersen_statistical_2012}. Logarithmic difference between values $V_1$ and $V_2$ is defined as $\ln(V_2)-\ln(V_1) = \ln(V_2/V_1)$. This is sometimes also referred to as log percent or L\% when the result is multiplied by 100 \parencites{tornqvist_how_1985,wetherell_log_1986}, logarithmic growth rate \parencite{casler_why_2015}, log points, nepers (centinepers in the case of multiplication with 100), decibels (when using $\log_{10}$), or logarithmic growth rates.
	Measuring change on a logarithmic scale has three useful related advantages over the often used percentage change, defined as $(V_2-V_1)/V_1\cdot 100$. These are symmetry, additivity, and the lack of extreme positive outliers. 
	
	The absolute value of log change between two counts is the same regardless of which is used as the reference point. Given a series of log changes, the final (log) frequency is equal to the sum of the initial (log) frequency and the series of log changes. Percentage change is by definition bounded at -100\% on the negative end, while increases starting at small values yield very large positive numbers. 
	
	Log change has the disadvantage that any 0-counts must be smoothed to avoid negative infinity resulting from $\ln(0)$, while for percent change, smoothing is strictly necessary only for increases from 0 to non-0 (to avoid division by 0), as a decrease from non-0 to 0 is always -100\% (regardless of the actual difference between the two values, which in itself may be seen as another disadvantage, depending on the use case).
	Simple $+1$ smoothing could be used to avoid this problem by incrementing all frequencies by 1. This leads to some bias when dealing with relatively small values (particularly after normalizing to per million words). We use a slightly more elaborate version where we only change any 0 values involved in frequency change calculations to the value that corresponds to 1 occurrence in the per-million normalized frequency counts, and leave all $>0$ values untouched. 
	
	Log frequencies are also better suited than raw frequencies (and absolute change) when dealing with word frequencies, smoothing the influence of the small number of extremely frequent words at the top end of the typically Zipfian distribution. We also tested the advection model using absolute frequency changes. Correlating absolute change based advection values with absolute frequency changes yields a practically zero correlation value. When using absolute frequencies for the advection calculation, but correlating these with log frequency changes, the correlations tend to come out as either the same or lower compared to using log change everywhere (as we do in this paper). In summary, there is little reason to not use log change to measure change. Table \ref{table:logchange} illustrates the differences of logarithmic and percentage measures of change in frequencies between two time periods, $t_1$ and $t_2$.
	
	
	\begin{table}[ht]
		\begin{center}
			\begin{tabular}{lrrrrr|rrrrr}
				$t_1$ & 1 & 5 & 50 & 1 & 10 & 10 & 10 & 100 & 100 & 100 \\ 
				$t_2$ & 10 & 10 & 100 & 100 & 100 & 1 & 5 & 50 & 1 & 10 \\ 
				\hline
				abs. change & 9 & 5 & 50 & 99 & 90 & -9 & -5 & -50 & -99 & -90 \\ 
				\% change & 900\% & 100\% & 100\% & 9900\% & 900\% & -90\% & -50\% & -50\% & -99\% & -90\% \\ 
				ln change & 2.3 & 0.69 & 0.69 & 4.61 & 2.3 & -2.3 & -0.69 & -0.69 & -4.61 & -2.3 \\ 
				$\log_{10}$ change & 1 & 0.3 & 0.3 & 2 & 1 & -1 & -0.3 & -0.3 & -2 & -1 \\ 
			\end{tabular}
		\end{center}
		\caption{
			Fictional word counts and the resulting change values using different measures. Note the asymmetry in percentage change values when the counts are flipped. Natural logarithms are rounded to save space.
		}
		\label{table:logchange} 
	\end{table}

	\subsubsection*{Additional remarks on using advection for time series adjustment}
	
	Table \ref{table:timeseries} illustrates the word frequency time series adjustment operation based the topical advection measure, described in Section \ref{results-decomposition}. The alphabetic abbreviations in the following equations refer to the rows in Table \ref{table:timeseries}. The decomposition-like adjustment is additive: the adjusted log change values $x = c - d$. 
	The frequency series can be reformed as the exponential of the cumulative sum of the adjusted values, initiated with the log frequency at period 1, $a_1$: 
	$$ y_i = e^{a_1 + \sum_{j=1}^{j=i} x_j } $$    
	
	This could be useful for visualization purposes, as on Fig.~\ref{fig:timeserieswords4}, but of course the actual values in the reformed series depend on the (arbitrary) initialization value. The values in the resulting reformed (exponentiated) series will never be negative, but may be very small, if topical advection for a given word at a given time point is considerably higher than its frequency change (we observe this to be rarely the case).
	
	\begin{table}[ht]
		\centering
		\begin{tabular}{l|rrrrrrrrrrr}
			& 1900s & 1910s & 1920s & 1930s & 1940s \\ 
			\hline
			(a) pmw frequency & 69.2 & 71.2 & 151.5 & 226.3 & 118.3 \\ 
			(b) log freq & 4.25 & 4.28 & 5.03 & 5.43 & 4.78 \\ 
			(c) log change &  & +0.03 & +0.75 & +0.4 & -0.64 \\ 
			(d) advection &  & -0.06 & +0.45 & +0.3 & -0.42 \\ 
			(x) adjusted log change &  & +0.09 & +0.3 & +0.1 & -0.23 \\ 
			(y) reformed series & 69.19 & 75.53 & 102.08 & 112.99 & 90.15 \\ 
		\end{tabular}
		\caption{
			Time series decomposition using topical advection on the example of the word \textit{payment}, corresponding to Fig.~\ref{fig:timeserieswords4}. Frequencies (a) are per million words. Log frequency and log change (b, c) refer to natural logarithms. The advection values (d) are based on the PPMI model with corpus topic smoothing. All values are rounded to save space and are therefore not precise.
			The increases in frequency of \textit{payment} in the 1920s and 1930s, as well as the decrease in the 1940s (cf. row c) coincide with the changes in the averaged frequency of the topic words of \textit{payment}, i.e., topical advection (d). The adjusted log change values (x) reflect the estimated frequency changes of \textit{payment} when topical fluctuations are accounted for. 
		}\label{table:timeseries}
	\end{table}

	\subsubsection*{Time series adjustment does not hide genuine competition}

	This section further supplement Section \ref{results-decomposition}, detailing the artificial corpus construction. The artificial series were inspected to see if the adjustment operation might possibly hinder the detection of actual competition between linguistic elements.
	
	We selected four test nouns of various frequencies that each: occur frequently enough in the corpus during the past century to evaluate their topics; exhibit relative stability across the 11 time periods (1900s-2000s) in terms of their occurrence frequency, as well as meaning (based on the APSyn measure (cf.\ Section~\ref{modelsection-previouswork}) on their context word vectors); and have small (absolute) advection values.  The words \textit{roof} (frequency at period 1: 163 per million words), \textit{reason} (724), \textit{town} (748), and \textit{face} (1938) satisfied these criteria. 
	
	We then generated artificial competing synonyms by replacing a linearly-increasing proportion of the occurrences of each of the four target words with an invented ``synonym" (\textit{word$'$}) in the corpus.
	We also experimented with an S-shaped increase curve \parencite[arguably more characteristic of language change, cf.][]{blythe_s-curves_2012}, which did not change the results.
	For example, at period 1, the invented synonym \textit{town$'$} appears nowhere in the manipulated corpus, while in period 2, 10\% of the occurrences of \textit{town} are  replaced with \textit{town}$'$ in the manipulated corpus, 20\% in period 2 and so on up to $100\%$ in period 11. Importantly, the replacement positions in the corpus were sampled at random, in order to simulate a scenario where the two synonyms are used freely (i.e., without regard for any contextual factors like style or genre).
	
	On applying the advection correction to each of the original words and their synonyms, we find their frequency change points are only shifted slightly from their known values. When looking at the advection-adjusted fraction of occurrences of a word or its invented synonym (i.e., relative frequencies), the shifts due to the advection adjustment are barely noticeable. In other words, we find that advection-based adjustment does not seem to obscure genuine (although in this case artificial) cases of selection.

	\subsection*{Details on correlating advection model power and genre divergence}

	As mentioned in Section \ref{results-coha}, we found that the advection measure correlates positively with divergences between genre distributions in COHA. Data in the decade subcorpora in COHA is subsequently divided into four genres (fiction, magazines, news, non-fiction). We measured the genre distribution of each decade by counting the total number of words in each genre. 
	Genre distributions of successive decades were compared using Kullback-Leibler divergence (to avoid zeros in the calculation, we incremented zero word counts by 1, in the early decades lacking the ``news" genre). A value of 0 would indicate an identical distribution. The distribution of the aforementioned genres in the 1950s subcorpus is 50\%, 24\%, 14\% and 12\%. The difference to the 1940s is less than 1 percentage point in each genre, yielding a divergence of $0.00002$. The largest observed divergence value is $0.13$, between 1810s and 1820s, where ``magazines" and ``non-fiction" both differ by about 16 percentage points.
	
\begin{figure}[tbh]
		\noindent
		\includegraphics[width=\columnwidth]{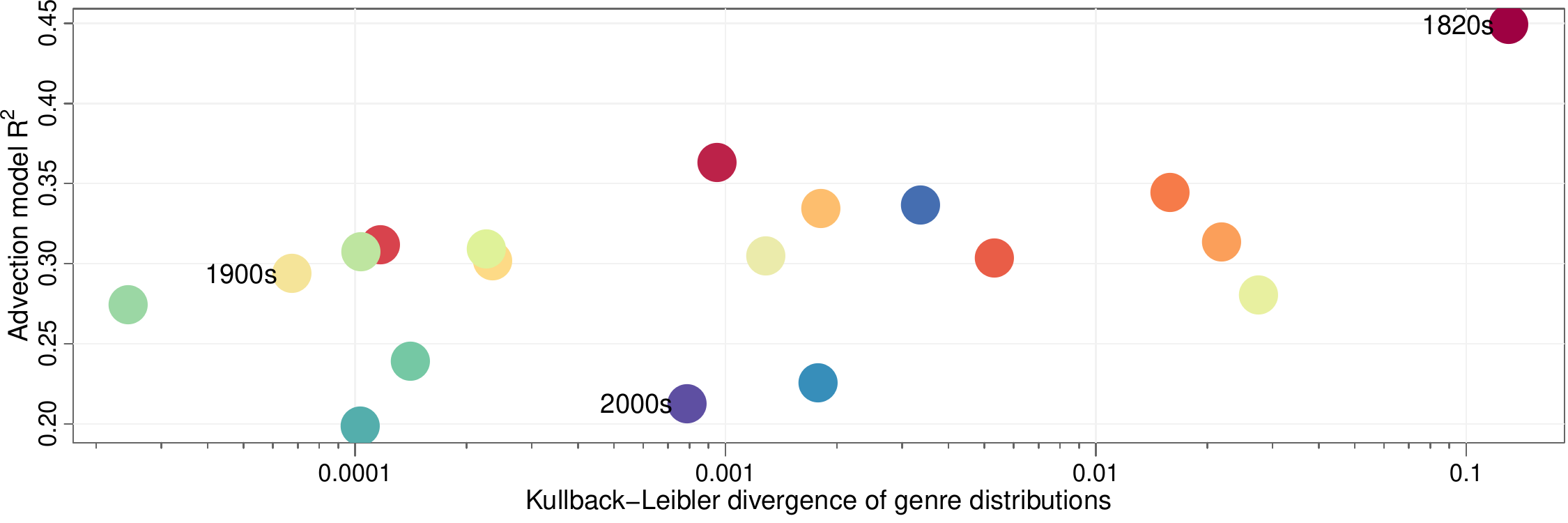}
		\caption{
			Divergence of genre distributions and the descriptive power of the advection measure (in the PPMI-based model, with smoothing). Each dot stands for one decade pair comparison, e.g. the dark purple dot marks the comparison of the 2000s to the preceding 1990s. The colors correspond to the colors in Fig. \ref{fig:scatter_with_examples}. Note the log scale on the horizontal axis. 
			Decade pairs where the advection model describes more variance in noun frequency changes tend to be the ones with higher divergence in genre distributions.
		}\label{fig:appendix_kullback}
	\end{figure}

	We find that (the log of) these divergence values correlates positively with the coefficients of determination from the advection model (i.e., the models where advection values are correlated with the word frequency change values). The $R^2$ values from correlating the divergence values to the $R^2$ values from the PPMI-based model without and with smoothing, and the LDA-based ones, without and with smoothing, in that order, are: 0.17, 0.41, 0.05, and 0.26.
	This indicates that the advection model is picking up on the changes between genre sample sizes, but also that discrepancies in genre sampling are likely not the only thing driving the observed changes in COHA over time. Figure \ref{fig:appendix_kullback} visualizes these results.
	
	\subsection*{Choice of corpora and methods, and their limitations}
	
	We used fairly large corpora --- COHA and COCA --- for our analyses, both of which have been described as relatively representative and well balanced in terms of genre. We excluded the first decades of COHA in some cases, due to their smaller size and less balanced nature. Notably, the ``news" genre is entirely missing in the first five decades. Mileage of utilizing the advection model with smaller corpora would probably vary, and is of course open for experimentation in terms of the parameters, thresholds and possibly the topical-semantic smoothing as described above. 
	It is not impossible that superior results could be potentially achieved using larger and better balanced corpora and more sophisticated methods of topic modeling with carefully optimized parameterizations (for example, our exploration of the LDA parameter space was admittedly fairly limited).

	\subsubsection*{Variations in operationalizing the test corpora}

	The results in Section \ref{results-coha} were based on comparing frequency changes between decade-length bins of the COHA. We also experimented with different temporal distances to see if the model behaves considerably differently. We found that with increased distance between the target decade and future decades, the values do improve in the case of some decade subcorpora, but not all, presumably depending on how much the subcorpora differ in terms of their underlying topic distribution. For example, the advection model describes more variance between mid-20th century decades and the 2000s compared to their immediate successors, while the 1810s subcorpus, clearly divergent in its distribution of genres and topics, shows relatively high correlations with all other subcorpora.

	We also experimented with applying the advection model to a shuffled corpus to test if there the observed correlation between word frequency changes and topical advection (cf. Section \ref{results-coha}) could be the result of some overlooked artifact of the model. We used the last decade subcorpus of COHA, but randomized the position of every word in the corpus, and calculated the topical advection value for all the target words, i.e. the weighted mean log context change (PPMI based, without smoothing), but using the randomized contexts. This resulted in $R^2 < 0.001$, $p=0.4$, indicating that the topical advection measure --- if calculated based on natural language use and not on random sequences of words --- does yield meaningful information about the frequency change in the topic of a word.

	\subsubsection*{Semantics, semantic change, and corpus smoothing}
	
	We re-evaluated the topics of words for every period to accommodate for natural semantic change. In principle this may not be necessary, if the meaning of a word is known to be very stable across time. In this case, the context vector from a single period, or aggregated across periods, could be used. The latter would also remedy the inherent problem of inferring context vectors for low-frequency words.%
	
	We note that the advection model should not be affected by the recent critique of distributed semantics by \textcite{dubossarsky_outta_2017}, who show that semantic change measures based on vector spaces tend to be biased by differences in frequency. In particular, they call into question the entire enterprise of automatically measuring meaning change, attempting to replicate previous studies \parencites{dubossarsky_bottom_2015,hamilton_diachronic_2016} and finding that the proposed results either do not hold up or have drastically diminished descriptive power in comparisons against randomized baselines --- attributing them to problems in vector space construction methods as well as bias from word frequency. 
	
	The same context word vectors we use to determine topic could indeed easily also be used to determine semantic change, by comparing the lists of top context words (cf. Fig.~\ref{fig:newwords}) between periods either by directly using the APSyn measure (cf. Section \ref{modelsection-previouswork}), or comparing the entire (suitably aligned) PPMI context vectors using vector cosine (in case of the former, care should be taken not to include 0-weight words in the topics, since APSyn only considers the rankings of context words in the vector, not their weights). 
	
	However, advection (topic frequency change) is meant to be re-evaluated for each corpus period. As such, semantic change is not directly a concern. We did also demonstrate additional results using what we called ``smoothing" (Section \ref{results}), or concatenating the data from the target period $t$ and the preceding period $t-1$ for the purpose of inferring topic vectors. In our experiments, this improved the power of advection to predict frequency change. In principle, smoothing could be applied using any number of $t\pm n$ periods; we also experimented with concatenating the entire corpus, and found that the descriptive power of the advection model suffered considerably. We assume semantic change to be the reason, since the context words (using which the advection measure is calculated) relevant to a target in one period may be quite irrelevant from another period, if the use (meaning) of the target differs --- leading to uninformative topics.
	
	Notably, the advection model is not expected to work as well with highly polysemous or general words (and homonyms), as it would with words with a more specific meaning (unless the meanings are somehow disambiguated and sense-tagged).
	The same goes for phrases and multi-word units, which we do not attempt to detect or parse in this contribution.
	Polysemy and multi-word units, however, are widespread problems across most NLP tasks, not only the one at hand.

	\subsection*{Notes on implementation}
	
	The models and calculations presented in this paper were implemented using R 3.5.0 \parencite{r_core_team_r_2018}, and making use of the \texttt{text2vec} package \parencite{selivanov_text2vec_2018}. The code and data are available at 
	{\scriptsize \url{https://github.com/andreskarjus/topical_cultural_advection_model}}.
	The corpora used here can be found at \url{https://corpus.byu.edu}.
	
	
	\printbibliography	
	
\end{document}